%% file: main.tex
\documentclass[journal]{IEEEtran}
\usepackage{graphicx}
\usepackage{amsmath}
\usepackage{amssymb}
\usepackage{booktabs}
\usepackage{multirow}
\usepackage{multicol}
\usepackage{mdframed}
\usepackage[skip=1pt]{caption}
\usepackage{lipsum}
\usepackage{algpseudocode}
\usepackage{algorithm}
\usepackage[pagebackref,breaklinks,colorlinks]{hyperref}
\usepackage{tikz}
\newcommand{\tikzxmark}{%
\tikz[scale=0.23] {
    \draw[line width=0.7,line cap=round] (0,0) to [bend left=6] (1,1);
    \draw[line width=0.7,line cap=round] (0.2,0.95) to [bend right=3] (0.8,0.05);
}}
\newcommand{\tikzcmark}{%
\tikz[scale=0.23] {
    \draw[line width=0.7,line cap=round] (0.25,0) to [bend left=10] (1,1);
    \draw[line width=0.8,line cap=round] (0,0.35) to [bend right=1] (0.23,0);
}}
\usepackage{subcaption}
\begin{document}
\input{sec/0_metadata}
\input{sec/0_abstract}
\input{sec/1_introduction}
\input{sec/2_related}
\input{sec/3_method}
\input{sec/4_results}
\input{sec/5_conclusions}
{
    \bibliographystyle{IEEEtran}
    \bibliography{main}
}

\input{sec/X_supplementary}

\end{document}

%% file: sec/0_metadata.tex
\title{Zero-Shot Machine Unlearning}

\makeatletter
\newcommand{\printfnsymbol}[1]{%
  \textsuperscript{\@fnsymbol{#1}}%
}
\makeatother

\author{Vikram~S~Chundawat\textsuperscript{$\dagger$}, Ayush~K~Tarun\textsuperscript{$\dagger$}, Murari~Mandal{$\ddagger$}, Mohan~Kankanhalli
\thanks{\textsuperscript{$\dagger$}Equal Contribution. The work is part of the authors' internship at the School of Computing, National University of Singapore.} 
\thanks{\textsuperscript{$\ddagger$}\textit{Corresponding Author}. Work performed while at the School of Computing, National University of Singapore.}
\thanks{Ayush~K~Tarun, Vikram~S~Chundawat are with Mavvex Labs, India, Faridabad 121001 (ayushtarun210@gmail.com; vikram2000b@gmail.com), Murari~Mandal is with the School of Computer Engineering, Kalinga Institute of Industrial Technology (KIIT) Bhubaneswar, India 751024 (Email: murari.mandalfcs@kiit.ac.in), Mohan~Kankanhalli is with the School of Computing, National University of Singapore (NUS), Singapore 117417 (Email: mohan@comp.nus.edu.sg)}
}
\maketitle

%% file: sec/0_abstract.tex
\begin{abstract}


Modern privacy regulations grant citizens the right to be forgotten by products, services and companies. In case of machine learning (ML) applications, this necessitates deletion of data not only from storage archives but also from ML models. Due to an increasing need for regulatory compliance required for ML applications,~\textit{machine unlearning} is becoming an emerging research problem. The right to be forgotten requests come in the form of removal of a certain set or class of data from the already trained ML model. Practical considerations preclude retraining of the model from scratch after discarding the deleted data. The few existing studies use either the whole training data, or a subset of training data, or some metadata stored during training to update the model weights for unlearning. However, strict regulatory compliance requires time-bound deletion of data. Thus, in many cases, no data related to the training process or training samples may be accessible even for the unlearning purpose. We therefore ask the question: \textit{is it possible to achieve unlearning with zero training samples?} In this paper, we introduce the novel problem of \textit{zero-shot machine unlearning} that caters for the extreme but practical scenario where zero original data samples are available for use. We then propose two novel solutions for \textit{zero-shot machine unlearning} based on (a) error minimizing-maximizing noise and (b) gated knowledge transfer. 
These methods remove the information of the forget data from the model while maintaining the model efficacy on the retain data. The zero-shot approach offers good protection against the model inversion attacks and membership inference attacks. We introduce a new evaluation metric,~\textit{Anamnesis Index} (AIN) to effectively measure the quality of the unlearning method. The experiments show promising results for unlearning in deep learning models on benchmark vision data-sets. The source code is available here: \url{https://github.com/ayu987/zero-shot-unlearning}
\end{abstract}
\IEEEpeerreviewmaketitle

%% file: sec/1_introduction.tex
\section{Introduction}
\label{sec:intro}
Forgetting or deletion of data may be desired due to different reasons such as request-to-delete from the data owner, inadvertent use of wrong data, and change in the data privacy rules. The recent introduction of data privacy and protection regulations (European Union's GDPR~\cite{voigt2017eu}, and California Consumer Privacy Act (CCPA)~\cite{goldman2020introduction}) obligate the companies/organizations to implement such deletion-upon-request framework. The data or anything derived from that particular data must be deleted. Machine learning (ML) applications are built upon the algorithms that observe and learn the information from the training data. Such ML models usually memorize the training data~\cite{feldman2020does,carlini2019secret}. Machine unlearning refers to the task of making a trained machine learning model forget a cohort or class of data. The objective here is to forget a certain class of data without affecting the performance on the remaining class of data, i.e. preserving the accuracy of the ML model for the remaining classes. Moreover, the unlearned model must not reveal information about the forget data i.e., it must be robust to privacy attacks. Several studies~\cite{golatkar2020eternal,bourtoule2021machine} have shown this to be a non-trivial problem. Full retraining of the ML model for every instance of such request would lead to huge computational costs. It is crucial to accomplish unlearning in a time and cost efficient manner. The class-level unlearning is useful in many application like face recognition and even healthcare applications. A Face ID is treated as a class in the face recognition model. Similarly in healthcare applications, a patient may ask for removal of her data from the  already trained ML model. The user-level data may be removed with the help of an unlearning method without retraining the entire model from scratch.\par

Recently, machine unlearning has received considerable attention~\cite{golatkar2020eternal,wu2020deltagrad,golatkar2020forgetting,golatkar2021mixed,tarun2021fast,tarun2022deep,guo2020certified}. Most of the solutions are limited to simple linear and logistic regression. Few methods~\cite{golatkar2020eternal,golatkar2020forgetting,golatkar2021mixed} have been proposed that can forget information from the CNN network weights with reasonable success in small to large scale vision problems. Unlearning in a CNN is quite difficult due to its vastly non-convex loss-landscape which makes it difficult to model the effect of a data sample on the optimization trajectory and the final network weights configuration.~\cite{tarun2021fast} proposed an efficient, multi-class unlearning algorithm that works across variety of deep networks such as the CNN and vision transformers. All these methods require~\textit{access to the retain data~\cite{golatkar2020eternal}}, or the \textit{forgetting data~\cite{guo2020certified}}, or sometimes both. However,~\textit{none of them can function when no original training data is available}. 

\textbf{Our Contribution.}
Machine learning (ML) models are frequently trained with data gathered from different sources such as publicly available datasets, data marketplaces, and self-collected/created data. In some cases, personal data such as medical records and facial images are collected with the consent of an individual. Many organizations \footnote{\href{https://cloud.google.com/products/ai}{https://cloud.google.com/products/ai}}
\footnote{\href{https://aws.amazon.com/machine-learning/}{https://aws.amazon.com/machine-learning/}}
\footnote{\href{https://azure.microsoft.com/services/machine-learning/}{https://azure.microsoft.com/services/machine-learning/}} also offer a limited duration access to certain datasets and facilitate ML model training in the cloud. The complete access to the dataset is never desired as in most cases the company uses only the trained model in their applications. Moreover, the data is usually chargeable for each instance of access. Therefore, multiple access to the data for unlearning purpose will be expensive. On the other hand the public datasets may not be available for reuse at later points in time and storing a copy of all the training data is expensive and impractical in many cases. All these factors suggest that machine unlearning algorithms that rely on the availability of original training data will not be helpful in many practical scenarios. The existing machine unlearning methods~\cite{golatkar2020eternal,wu2020deltagrad,golatkar2020forgetting,golatkar2021mixed,tarun2021fast,guo2020certified} would fail to work without having access to at-least some subset of the original training data. This motivated us to ask the following question: \textit{Can we develop machine unlearning algorithms that require no training samples or information related to the training process?} The algorithms offering solution to this question would be categorized as \textit{zero-shot machine unlearning algorithms}.\par

While building a ML system based on a diverse set of data, a company also needs to comply with the data privacy rules and regulations.
The data privacy regulations~\cite{voigt2017eu,goldman2020introduction} are evolving over time to include the \textit{right to be forgotten}. They are likely to include a variety of data-usage restrictions in future to give individuals more control over their data for the sake of their privacy. For example, the CCPA~\cite{goldman2020introduction} permits companies to collect user data without consent but compels them to remove the user's data upon a request to opt-out of data sharing. If the original data or some portion of the data is lost, then the existing machine unlearning methods will not be of any help. In such a situation the entire ML model might have to be deleted to remain compliant with the regulations. This can be quite disruptive to the company's business. Furthermore, even if the data is available, unlearning is much more efficient as compared to retraining. In this paper, we present a data-free solution to machine unlearning that addresses the unique problem of zero data availability. In case of no access to the original training samples, the proposed zero-shot unlearning algorithms become an important tool for the company. With efficient machine unlearning, companies can ensure the compliance of their ML models with the relevant privacy rules and regulations.\par

~\textit{How would a machine unlearning algorithm function with zero training data?} In order to manipulate the network weights of the model for unlearning, some kind of stimulus is needed. One possible approach could be the model is shown certain patterns that represent the same feature distribution as the original data samples. These patterns could be generated through generative adversarial networks (GANs) or variational autoencoders (VAEs). It is to be noted that we don't know anything about the original training data distribution. Thus, generating relevant samples would be tricky through GANs/VAEs. Another possible approach could be to build upon the unlearning algorithm presented in~\cite{tarun2021fast} which doesn't require access to the forget data. Similar to the error maximizing noise generated for the forget class(es) in~\cite{tarun2021fast}, error minimizing noise can be generated for the remaining classes. These learned noise matrices could be used to update the model for unlearning. Yet another approach could be to transfer the knowledge of certain class(es) by filtering out their information into another network in a teacher-student framework. Motivated by the aforementioned propositions, we present two different methods for achieving zero-shot machine unlearning.

In this paper, we formally define the problem of zero-shot unlearning and make the following key contributions:

\begin{itemize}
    \item We introduce the problem setting for zero-shot machine unlearning that imposes the constraint that \textit{zero training data} is available to the unlearning algorithms. This resembles the real-world scenarios more closely for data deletion request in ML models as compared to the existing state-of-the-art machine unlearning settings. 
    
    \item We propose two novel methods to enable data-free unlearning. These methods are based on (i) error-maximizing and error-minimizing noise and (ii) gated knowledge transfer in a teacher-student learning framework.
    
    \item We introduce a new metric~\textit{Anamnesis Index}, to assess the quality of unlearning more effectively. We also evaluate our unlearned model against the privacy attacks such as model inversion attack and membership inference attacks. 
    
    \item  The proposed methods are independent of the prior information (such as the optimization technique) related to the original model training. We obtain strong performance on a variety of deep networks on benchmark vision datasets (MNIST, SVHN, CIFAR-10). 
\end{itemize}

%% file: sec/2_related.tex
\section{Related Work}
\label{sec:related}
\subsection{Machine Unlearning} 
The concept of unlearning was introduced in~\cite{cao2015towards} to eliminate the effect of data point(s) on the already trained model. Machine unlearning can be broadly grouped into two categories: \textit{exact unlearning and approximate unlearning methods.} In \textit{exact unlearning}, the impact of the data point (to forget) is removed from the model by retraining the model from scratch by excluding that data from the training set. This is clearly a computationally expensive process as the deep learning models are trained with large datasets. Moreover the request for data deletion is a recurrent event rather than a one-time occurrence. Several research efforts have focused on finding efficient ways of retraining.~\cite{bourtoule2021machine} proposed a SISA approach to train the model by partitioning the dataset into a set of non-overlapping shards. This reduces the need for full retraining as the model can be retrained on one of the shards. The~\textit{approximate unlearning} methods aim to approximate the parameters that would have been obtained if the model was trained without using the data to be unlearned. Usually the approach is to bring the parameters closer to a model (trained from scratch) that has never seen the unlearning data through a lower number of updates in comparison to exact unlearning methods.~\cite{graves2021amnesiac} store the updates made by each data point in the parameter space while training. During unlearning, the corresponding updates are subtracted from the final parameters of the model.~\cite{brophy2021machine} proposed random forests that support data forgetting.~\cite{ginart2019making,mirzasoleiman2017deletion} studied data deletion in k-means clustering algorithms. Similarly, unlearning for Bayesian methods are presented in~\cite{nguyen2020variational}.~\cite{guo2020certified} give a certified information removal framework based on Newton's update removal mechanism for convex learning problems.~\cite{mahadevan2021certifiable} studied several unlearning methods for linear models and analyzed the efficiency, effectiveness and certifiability trade-offs among them. Similarly,~\cite{sekhari2021remember} analyze the difference between machine unlearning and differential privacy. All these methods are designed for convex problems, whereas we aim to present an unlearning solution for deep learning models. 

\subsection{Deep Machine Unlearning}
Golatkar et al.~\cite{golatkar2020eternal} proposed an information theoretic method to scrub the information from intermediate layers of deep networks trained with SGD.~\cite{golatkar2020forgetting} extended this work to update the final activations of the model. They present a neural tangent kernel (NTK) based approximation of the training process and use it to estimate the updated network weights after unlearning. However,  the approximation accuracy decreases and computational cost increases very quickly even for small datasets.~\cite{golatkar2021mixed} directly train a linearized network and use it for unlearning. They train two separate networks: the core model, and a mixed-linear model for unlearning purpose. However, designing a mixed-linear network for every deep architecture is an inefficient approach and requires manual intervention depending on the network structure of the original model. In our zero-shot unlearning methods, we do not train any additional network. In fact, we do not require any prior information related to the training process or access to the training data. Tarun et al.~\cite{tarun2021fast} proposed an error-maximization based method to learn noise matrix for the class to forget. They use such anti-samples to induce class-level forgetting in deep networks. The method requires a small subset of retain data for unlearning. In this paper we work in a setting where strictly no information about the training data is available.\par

\subsection{Unlearning Settings and Data Privacy} The~\textit{approximate unlearning} methods aim to offer higher efficiency by making certain assumptions about the availability of training information i.e., training data, optimization techniques used, and relaxing the effectiveness of the model to some degree. Most of the existing unlearning methods can be provisionally categorized into three groups. The~\textit{first group}~\cite{guo2020certified,izzo2021approximate} use the data to forget to update the ML model for unlearning. In these methods, the influence of the forget data on the model is approximated with a Newton step and a random noise is injected to the training objective function. The~\textit{second group}~\cite{golatkar2020eternal,golatkar2020forgetting,golatkar2021mixed} requires access to the rest of the training data (excluding the forget data). These methods use the Fisher information and inject optimal noise to the model weights to achieve unlearning. They do however impose certain restrictions on the training process such as only SGD optimization is allowed while training.~\cite{tarun2021fast} alleviated some of these restrictions while presenting a much faster unlearning algorithm. The~\textit{third group}~\cite{graves2021amnesiac,neel2021descent,wu2020deltagrad,wu2020priu} stores different kind of information during the training process in order to utilize it for data deletion in future. Usually, these methods attempt to approximate the SGD steps of a model retrained from scratch. To this end, the intermediate elements such as the loss gradients, weight updates, etc. generated at each SGD step is stored. The amount of information that needs to be stored raise the concern of excessive memory overheads requiring a trade-off between memory overhead vs computational efficiency.\par

All the above mentioned methods require access to either the data to unlearn, the remaining data, or some metadata stored during training. These assumptions do not reflect the real-world unlearning scenarios as discussed in Section~\ref{sec:intro}. With increasing data privacy concerns~\cite{voigt2017eu,goldman2020introduction} in different parts of the world, more stringent privacy regulations are expected. Thus, a data-free approach to unlearning is required. The possible information leakage in the existing machine unlearning methods as discussed in~\cite{chen2021machine} also motivated us to look towards a completely data-free approach to unlearning. We propose to work in the zero-shot setting where the proposed unlearning method does not require access to any kind of information that are essential for the functioning of existing state-of-the-art unlearning methods. A data-free approach makes this a more realistic setting for unlearning and also helps in complying with a broader set of data privacy regulations.\par

\subsection{Privacy Attacks on Machine Learning Models}
Numerous privacy attacks against ML model have shown information leakage of the training data~\cite{shokri2017membership,salem2019ml}. Membership inference attack aims to extract the information about the presence of certain data points the training set~\cite{yeom2018privacy,dwork2015robust}. Shokri et al.~\cite{shokri2017membership} proposed the idea of using shadow models to imitate the behaviour of the target model to generate training samples. Thereafter, several membership inference attacks have been proposed~\cite{nasr2019comprehensive}. To safeguard a ML model from the inference attacks, considerable defense methods have been proposed based on reducing the model overfitting~\cite{li2020membership}, perturbation of the posteriors~\cite{jia2019memguard}, and adversarial training~\cite{nasr2018machine}. Ganju et al.~\cite{ganju2018property} introduced a subcategory of membership inference attack called property inference attack. It attempts to infer the general properties of the training data. For example, the ratio between the samples in each class is a property of the training data. Model inversion attacks~\cite{fredrikson2015model,hitaj2017deep} try to recreate instances of records of a class from the target ML model. Similarly, the adversarial examples~\cite{papernot2017practical,papernot2016limitations} are used to infer certain attributes of the data. In this paper, the membership inference attack and model inversion attack are used to check the possible privacy leaks in the unlearning models. 

%% file: sec/3_method.tex
\section{Preliminaries}
Let $\mathcal{D} = \{x_i, y_i\}_{i=1}^N$ be a dataset consisting of $x_i$ samples associated with label $y_i \in \{1,2,...,K\}$, each representing a class. $C_f$ denotes the set of classes we wish the machine learning (ML) model to forget and $\mathcal{D}_f$ is the set of data corresponding to the forget classes. Similarly, $C_r$ denotes the set of retain classes that we want the model to remember and $\mathcal{D}_r$ is the data corresponding to the retain classes.
A general machine unlearning method receives a query to forget $\mathcal{D}_f$ samples. The information about $\mathcal{D}_f$ data points are to be removed from the model. The usual assumption is that access to both $\mathcal{D}_r$ and $\mathcal{D}_f$ is available to the unlearning method. The unlearning may be performed at class-level or sample-level. In class-level unlearning, the entire class of data is forgotten. In sample-level unlearning, a particular set of data points either belonging to single or multiple classes are forgotten.\par

The complete dataset $\mathcal{D}$ consists of  $\mathcal{D}_f$ and $\mathcal{D}_r$ i.e., $\mathcal{D}_r\cup \mathcal{D}_f = \mathcal{D}$ where $\mathcal{D}_f$ and $\mathcal{D}_r$ are mutually exclusive i.e., $\mathcal{D}_r \cap \mathcal{D}_f = \phi$. The model trained from scratch without observing the forget samples ($\mathcal{D}_r$) is denoted as \textit{retrained model} or \textit{gold model} in this paper. In the literature, the retrained model is commonly used as a benchmark to evaluate the performance of an unlearning method. The Kullback-Leibler (KL) divergence~\cite{kullback1951information} is used as a measure of similarity between two probability distributions. Given two distributions $p(x)$ and $q(x)$, their KL-divergence is defined by
\begin{equation}
KL(p(x)||q(x)) := \mathbb{E}_{x\sim p(x)}[log(p(x)/q(x))]
\end{equation}
where $KL(p(x)||q(x))$ is always positive.

\section{Zero-Shot Machine Unlearning}
In this section we first formalize the \textit{zero-shot unlearning problem}. We then present two novel approaches for \textit{zero shot unlearning}. We also present a new evaluation metric to more effectively assess the degree of unlearning achieved in a model. 

\begin{figure}[]
\centering
\includegraphics[width=0.8\linewidth]{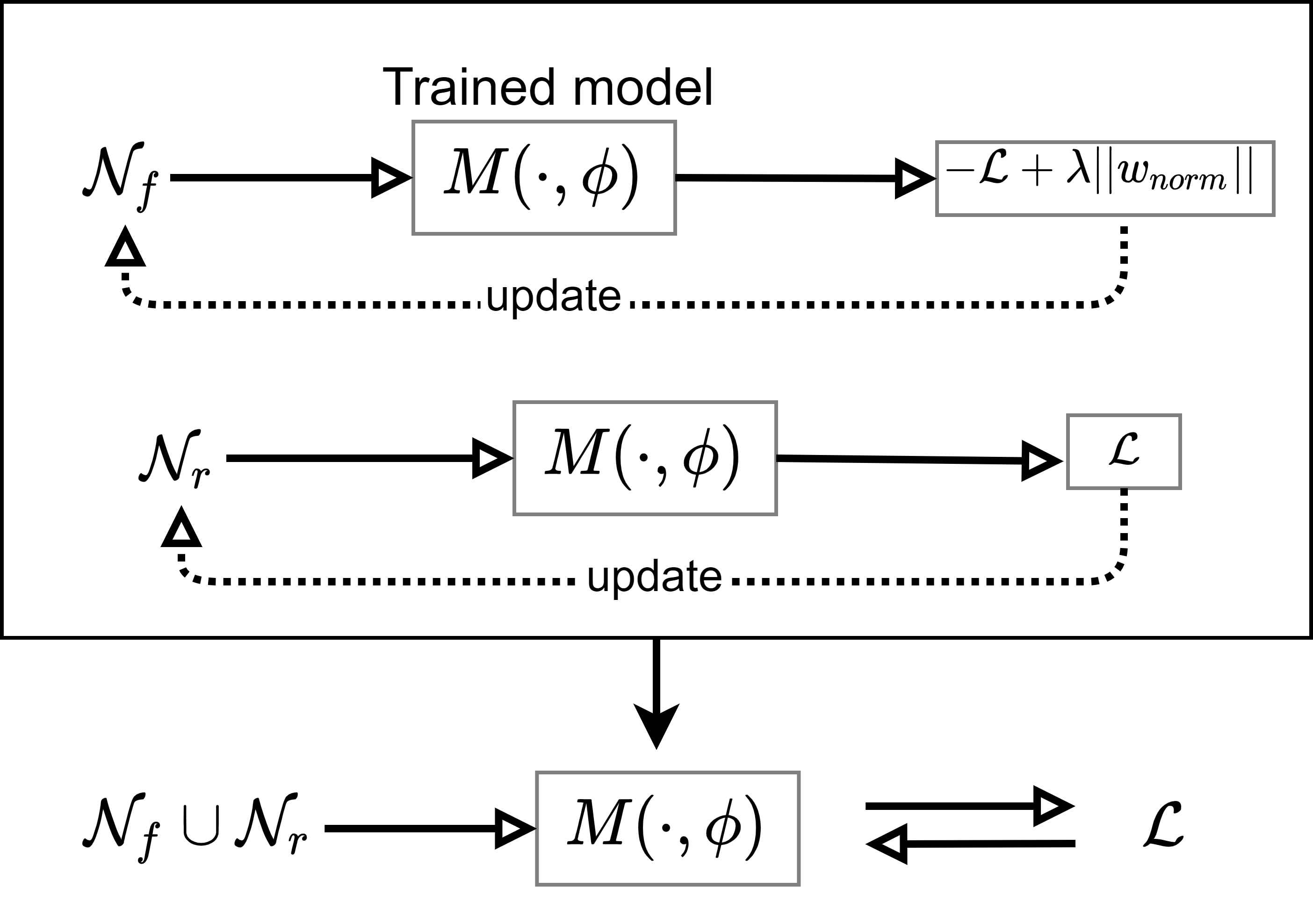}
\caption{The baseline error minimization-maximization noise based method for zero-shot unlearning.}
\label{fig:noise_framework}
\end{figure}

\subsection{The Zero-Shot Machine Unlearning Problem}
Let the machine learning (ML) model be represented by $M(x;\theta)$ that gives out the classification probabilities for each class, where $x$ is the input and $\theta$ is the set of model parameters. For an ML algorithm $A$ trained on the complete dataset $\mathcal{D}$, the model parameter set is obtained as 
\begin{equation}
\theta = A(\mathcal{D})
\end{equation}

If $M_{r}(x; \theta_r)$ is a model that has observed only the data corresponding to retain classes, then the parameters corresponding to $M_r$ can be expressed as
\begin{equation}
\theta_r = A(\mathcal{D}_r)
\end{equation}

A method $\mathcal{U}$ is called a \textit{zero-shot unlearning} method if it doesn't require access to either $\mathcal{D}_r$ or $\mathcal{D}_f$. It requires access only to the parameters ($\theta$) of the trained model which is always available. Based on the query set of forget classes ($C_f$) (not the data corresponding to these classes), a \textit{zero-shot unlearning} method can produce the unlearned model $M_{u}$ with parameters $\theta_u$ such that it is behaviourally similar to a~\textit{retrained model} $M_r$ as given in Eq.~\ref{eq:test1} and Eq.~\ref{eq:test2}.

\begin{equation}
\label{eq:test1}
M(x; \theta) \underset{\mathcal{U}}{\to} M_{u}(x; \theta_{u})
\end{equation}

\begin{equation}
\label{eq:test2}
M_u(x; \theta_u) \approx M_r(x; \theta_r)
\end{equation}

The output of the model $M_u$ are expected to be similar to that of $M_r$, as if $M_u$ has never seen the data corresponding to forget set $\mathcal{D}_{f}$. Note that the parameter set~$\theta_u$ is not expected to be exactly the same as $\theta_r$ due to many degrees of freedom of parameters. Two models consisting of different distribution of parameters could show similar behaviour. An analysis on this phenomena as presented in~\cite{tarun2021fast}. It is shown that the \textit{retrained model} $M_r$, may not be the only~\textit{gold standard}. A very different set of parameters could exhibit behaviour very similar to that of $M_r$. Therefore, in practice, we compare the black-box metrics of the unlearned model with the retrained model for robustness analysis. The rest of the analysis is done through privacy attacks. 

\subsection{Error Minimization-Maximization Noise} \label{error_min_max}
We build upon the work presented in~\cite{tarun2021fast} that utilizes an error-maximization based noise generation procedure. A set of noise matrices is learned for the forget class of data using the original model. These matrices are then used to fine-tune the fully trained model in order to unlearn the information corresponding to the forget classes ($C_{f}$). The generated matrices act as a set of anti-samples to the model for the forget class(es). Therefore, after observing them, it forgets about that class of data. A subset of the data corresponding to retain classes ($C_{r}$) are also shown to the model to preserve the information for the retain classes. Although, the forget class of data is not needed but it requires access to the retain data.\par

We create a novel extension of this approach in a data-free setting. The data corresponding to the $C_{r}$ is generated by an error-minimizing noise. The error-minimized noise should act as a proxy for the samples belonging to $C_{r}$. Since, the original samples from $C_{r}$ class would also have minimum possible model loss. Thus, learning noise matrices that minimize the model loss is a natural way to synthesize such data points. As the retain data $\mathcal{D}_{r}$ is not available for updating the model in a data-free setting, we create the retain data samples using the model itself through error-minimization. An error-minimizing noise $\mathcal{N}^{(i)}$ of the same size as that of the model input will be generated for every retain class $i$. The following loss function is used to learn the noise matrix.
\begin{equation}
    L_N^{(i)}(\mathcal{N}_{r}^{(i)}) = \mathcal{L}(M(\mathcal{N}_{r}^{(i)};\phi), i)
\end{equation}

where $\mathcal{L}$ is the classification loss. The noise is generated and updated for all the retain classes. The error-maximizing noise is learned in a manner similar to that in~\cite{tarun2021fast}. We do not use the repair step in our experiments, as it did not have its intended effect of improving the accuracy on the retain set in the data-free setting. Also, we do not restrict ourselves to only a single shot of impair and use  multiple iterations to obtain the best performance possible. The overall method is depicted in Figure~\ref{fig:noise_framework}. Furthermore, a step-wise procedure can be found in the supplementary material. 

\begin{figure}[]
\centering
\includegraphics[width=0.8\linewidth]{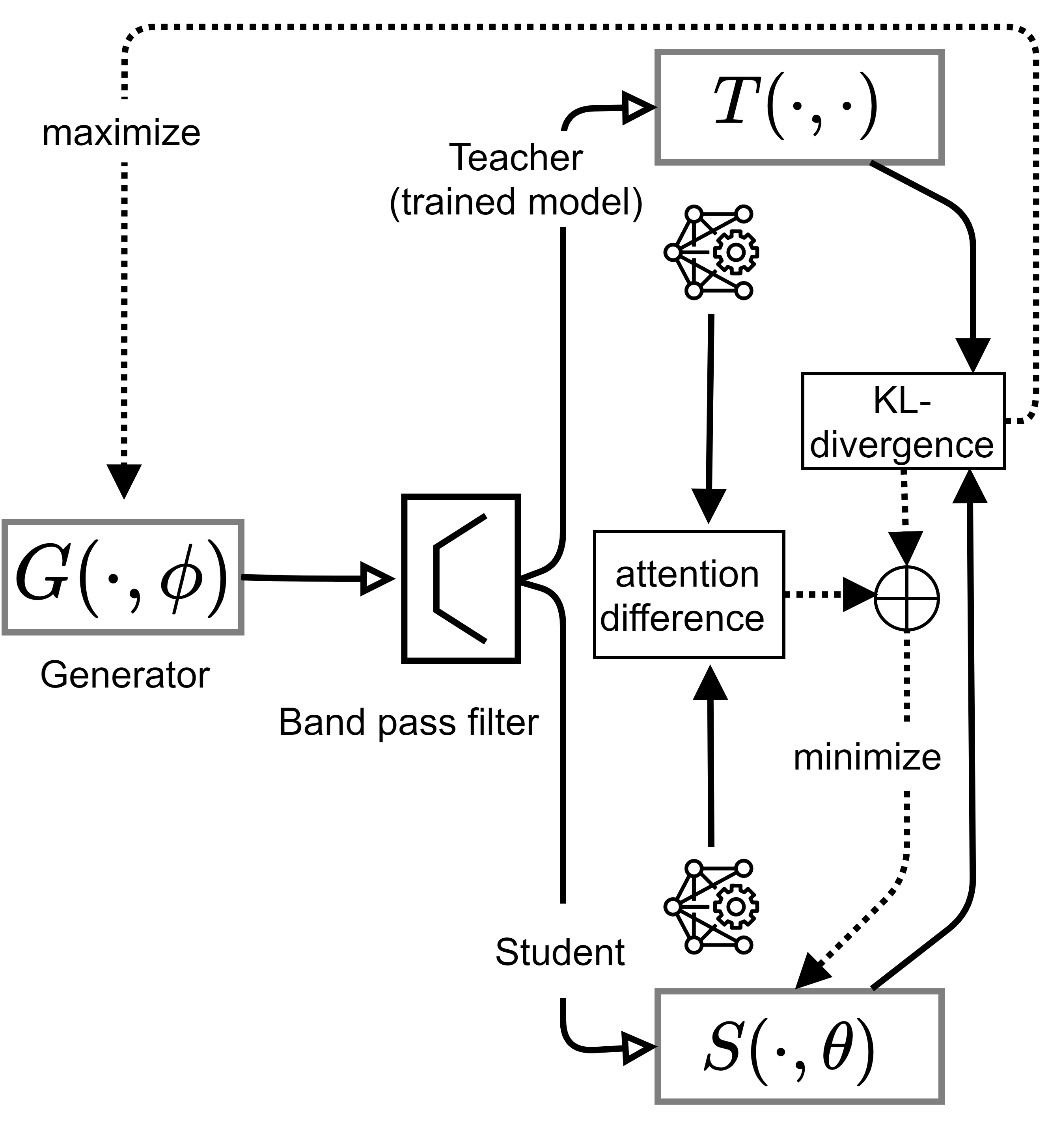}
\caption{The proposed gated knowledge transfer method for zero-shot unlearning.}
\label{fig:GKT_framework}
\end{figure}

\subsection{Gated Knowledge Transfer} 
\label{GKT_Section}
The sub-optimal quality of the learned noise generated through the error maximization-minimization method leads to substandard level of unlearning. We therefore propose a knowledge distillation based algorithm to obtain better performance for the data-free setup. Inspired by~\cite{micaelli2019zero}, we use a generator to produce data points that maximize the KL-divergence between the teacher and student models. However, instead of trying to obtain a smaller model, we develop a zero-shot mechanism for machine unlearning through such knowledge distillation. To this end, we assign the originally trained model as the teacher and a randomly initialized network (of same structure as the original model) as the student. We introduce a crucial design  element, a type of \textit{"gate"} we call \textit{band-pass filter} which does not allow the information related to the forget class $C_{f}$ to flow towards the student from the teacher model. This ensures that the student learns only the information of the retain class(es) $C_{r}$ and not the forget class. The proposed method is shown in Figure~\ref{fig:GKT_framework}.\par  

Let $T(x)$ be a teacher which outputs the probability vector $t$ for input $x$. The student $S(x; \theta)$ is a randomly initialized model with parameters $\theta$, output probabilities $s$, and has the same architecture as $T(x)$. The generator $G(z;\phi)$ produces pseudo data points $x_p$ for a noise vector $z \in N(0, I)$. Let the corresponding teacher's prediction on $x_p$ be $t_p$ and student's prediction on $x_p$ be $s_p$. The generator maximizes $D_{KL}(T(x_p) || S(x_p))$ i.e., Kullerback-Leibler(KL) divergence between output probabilities of the teacher and student model as shown in Eq.~\ref{eq_kl_divergence}.
\begin{equation}
\label{eq_kl_divergence}
    D_{KL}(T(x_p) || S(x_p)) = \sum_{i}{t_p^{(i)}log(t_p^{(i)}/s_p^{(i)})}
\end{equation}
where $i$ corresponds to the data classes. The data points $x_{p}$ are generated by the generator which is optimized to maximise the KL-Divergence between the teacher and student. The student will update the weights to minimize the KL-divergence along with an attention loss as given in Eq.~\ref{eq_kl_divergenc_2}.
\begin{equation}
\label{eq_kl_divergenc_2}
    L_{at} = \sum_{l \in N_L}^{}{\Bigg{|}\Bigg{|} \frac{f(A_l^{(t)})}{||f(A_l^{(t)})||_2}} - \frac{f(A_l^{(s)})}{||f(A_l^{(s)})||_2}\Bigg{|}\Bigg{|}
\end{equation}
where $A_l^{(t)}$ and $A_l^{(s)}$ denote outputs of layer $l$ for teacher and student respectively, both of which consist of $N_{A_l}$ channels. Similar to~\cite{micaelli2019zero}, we use a subset $N_L$ for calculating the attention loss. We use $f(A_l) = (1/N_{A_l})\sum_{c}{a_{lc}^2}$ where $a_{lc}$ denotes the $c_{th}$ channel of activation block $A_l$. The student then minimizes $L_s$ as in~Eq.~\ref{eq_kl_divergenc_3}.
\begin{equation}
\label{eq_kl_divergenc_3}
    L_{s} = D_{KL}(T(x_p) || S(x_p)) + \beta L_{at}
\end{equation}
where $\beta$ is a hyperparameter. The generator and student are updated alternatively.

\subsubsection{\textbf{Band-Pass Filter.}} We propose a band-pass filter which would attenuate the information regarding the forget class(es) $C_{f}$ and only let information regarding the retain classes $C_{r}$ to reach the student. The intuition is that as the samples aren't from a real world distribution, so the only indicator of how much information the generated pseudo samples have regarding the forget class(es) is the predicted probabilities by the teacher, $t_p^i$, where $i \in C_f$. To restrict the generator from passing on information  regarding the forget class(es), we place a filter $F$ ahead of the generator which receives all the generated pseudo samples and filters them out before passing them to the student. The filter criterion for each sample is 
\begin{equation}
\label{eq_filter}
    F(x_p) = \prod_{i \in C_f}{(t_p^{(i)} < \epsilon})
\end{equation}
where $\epsilon$ is a hyperparameter and $i$ represents each forget class. The filter creates a boolean mask and a sample passes the filter only if the predicted probability corresponding to each forget class i.e. $t_p^i$ is less than a threshold $\epsilon$ for all $i \in C_f$. Let the total number all of classes be denoted by $c$, then a randomly initialized model will on an average output a probability of $1/c$ for each of the forget class(es). Therefore, in order to effectively filter out the information regarding the forget class(es), a value of $\epsilon$ equal to or lower than $1/c$ would be required, but if $\epsilon$ is too small, no pseudo sample might pass the filter. Thus, an appropriate value of $\epsilon$ is crucial for better training. The step-wise procedure for the Gated Knowledge Transfer (GKT) method is presented in the supplementary material.

\begin{table*}[]
\footnotesize
\centering
\caption{Unlearning results in AllCNN model. \textbf{Original Model} denotes the model trained on the complete dataset. \textbf{Retrain Model} denotes the model trained from scratch only on $\mathcal{D}_{r}$. \textbf{M-M (Min-Max)} denotes the model generated by error minimization-error maximization technique. \textbf{GKT} denotes the gated knowledge transfer method. \textbf{AIN} denotes the Anamnesis Index. \# $\mathcal{Y}_{f}$ is the number of unlearning classes.}
\begin{tabular}{c|cccccc|ccc}
\hline
\multirow{2}{*}{Dataset} & \multirow{2}{*}{\# $\mathcal{Y}_{f}$} & \multirow{2}{*}{Acc.} & Original  & Retrain & M-M  & GKT & \textit{AIN} & \textit{AIN}\\
{} & {} & {} & Model & Model & Method & Method 
&[GKT]
&[M-M]\\
\hline
\multirow{4}{*}{CIFAR10} & \multirow{2}{*}{1} &  $\mathcal{D}_{r}$ $\uparrow$ & 84.05 & 85.72 &20.48 & \textbf{81.97} & \multirow{2}{*}{\textbf{0.81}} & \multirow{2}{*}{0.11} \\

{} & {} &  $\mathcal{D}_{f}$ $\downarrow$ & 87.49 & 0 & 5.11 & \textbf{0} & {} & {}\\
\cline{2-9}
 & \multirow{2}{*}{2} &  $\mathcal{D}_{r}$ $\uparrow$ & 84.18 & 86.30 & 29.59 & \textbf{81.70} & \multirow{2}{*}{\textbf{0.74}} & \multirow{2}{*}{0.10} \\
 
 {} & {} &  $\mathcal{D}_{f}$ $\downarrow$ & 84.72 & 0 & 7.59 & \textbf{0} & {} & {}\\
\hline

\multirow{4}{*}{SVHN} & \multirow{2}{*}{1} &  $\mathcal{D}_{r}$ $\uparrow$ & 94.52 & 93.02 & 72.92 & \textbf{92.43} & \multirow{2}{*}{\textbf{0.37}} & \multirow{2}{*}{0.15} \\

{} & {} &  $\mathcal{D}_{f}$ $\downarrow$ & 95.16 & 0 & 42.32 & \textbf{0} & {} & {} \\
\cline{2-9}

 & \multirow{2}{*}{2} &  $\mathcal{D}_{r}$ $\uparrow$ & 93.61 & 95.10 & 58.70 & \textbf{92.13} & \multirow{2}{*}{\textbf{0.74}} & \multirow{2}{*}{0.13}  \\

{} & {} &  $\mathcal{D}_{f}$ $\downarrow$ & 96.39 & 0 & 50.23 & \textbf{0} & {} & {} \\
\hline

\multirow{4}{*}{MNIST} & \multirow{2}{*}{1} &  $\mathcal{D}_{r}$ $\uparrow$ & 97.84 & 99.25 & 10.57 & \textbf{97.12} & \multirow{2}{*}{\textbf{0.65}} & \multirow{2}{*}{0.31} \\

{} & {} &  $\mathcal{D}_{f}$ $\downarrow$ & 99.61 & 0 &0.0 & \textbf{0} & {} & {} \\
\cline{2-9}
 & \multirow{2}{*}{2} &  $\mathcal{D}_{r}$ $\uparrow$ & 98.17 & 99.41 & 10.96 & \textbf{96.87} & \multirow{2}{*}{\textbf{0.30}} & \multirow{2}{*}{0.20} \\
{} & {} &  $\mathcal{D}_{f}$ $\downarrow$ & 98.77 & 0 & 0.0 & 0 & {} & {} \\
\hline
\end{tabular}
\label{table:allcnn}
\end{table*}

\subsection{Anamnesis Index}\label{AIN}
The existing methods~(\cite{tarun2021fast} \cite{golatkar2020eternal} \cite{golatkar2020forgetting}) use \textit{relearn time} as a metric to measure the quality of unlearning. It is used to represent the amount of information left in the models after unlearning by showing how quickly the model can relearn. However, the scale and properties of this metric could vary according to the model and the dataset. In our experiments, we noticed that sometimes the unlearned models regain significant accuracy in a very few number of relearn steps, but do not converge to the original accuracy on the forget class(es) for a large number of steps. Thus, simply calculating the retrain time (epochs) in which the original accuracy is reached or surpassed could be misleading. 
To address some of these issues, we use a margin of $\alpha\%$ around the original accuracy to calculate the relearn time. Let the number of mini-batches (steps) required by a model $M$ to come within $\alpha\%$ range of the accuracy of the original model $M_{orig}$ on the forget classes be $r_t(M, M_{orig}, \alpha)$. 
If $M_u$ and $M_s$ denote the unlearned model and the model trained from scratch on $\mathcal{D}_{r}$ respectively, then Anamnesis Index (AIN) is defined as
\begin{equation}
\label{eq:ain}
    AIN = \frac{r_t(M_u, M_{orig}, \alpha)}{r_t(M_{s}, M_{orig}, \alpha)}
\end{equation}
The value of AIN ranges from 0 to $\infty$. The closer the AIN value is to 1, the better is the unlearning. AIN values much lower than 1 correspond to the instances when information of the forget classes is still present in the model. It also indicates the unlearned model quickly relearns to make accurate predictions. This may be because there were only a few changes in the final layers that deteriorated the model's performance on forget class(es) but are easily reversible. If AIN is much higher than 1, this may indicate that the method leads to changes in parameters which are so significant that the unlearning itself is detectable  (Streisand effect). This may be because the model was thrown into a convergence hyperplane far-off from its initial position and is thus not able to regain earlier learnt information regarding the forget class(es).


\subsubsection{\textbf{Setting the value of $\alpha$.}} The suggested value of parameter $\alpha$ is 5-10$\%$. The $\alpha$ is a measure of the accepted margin of difference in performance for retrain time calculation of the models. A high value of $\alpha$ means a very high margin, and could lead to a misleading score, always near to 1. A much lower value of $\alpha$ can result in unstable results.

%% file: sec/4_results.tex
\begin{table*}[]
\footnotesize
\centering
\caption{Single class unlearning on LeNet and ResNet9}
\begin{tabular}{c|cccccc|ccc}
\hline
\multirow{2}{*}{Dataset} & \multirow{2}{*}{Model} & \multirow{2}{*}{Acc.} & Original  & Retrain & M-M & GKT & \textit{AIN} & \textit{AIN}\\
{} & {} & {} & Model & Model & Method & Method &[GKT] &[M-M]\\
\hline
\multirow{4}{*}{CIFAR10} & \multirow{2}{*}{LeNet} &  $\mathcal{D}_{r}$ $\uparrow$ & 59.80 & 62.93 & 55.32 & \textbf{41.32} & \multirow{2}{*}{\textbf{211}} & \multirow{2}{*}{24} \\

{} & {} &  $\mathcal{D}_{f}$ $\downarrow$ & 65.25 & 0 & 23.98 & \textbf{0} & {} & {}\\
\cline{2-9}
 & \multirow{2}{*}{ResNet9} &  $\mathcal{D}_{r}$ $\uparrow$ & 84.83 & 85.61 & 10.85 & \textbf{56.83} & \multirow{2}{*}{\textbf{212}} & \multirow{2}{*}{12}\\

{} & {} &  $\mathcal{D}_{f}$ $\downarrow$ & 88.50 & 0 & 0 & \textbf{0} & {} & {}\\
\hline
\multirow{4}{*}{SVHN} & \multirow{2}{*}{LeNet} &  $\mathcal{D}_{r}$ $\uparrow$ & 85.69 & 88.31 & 81.80 & \textbf{78.27} & \multirow{2}{*}{\textbf{1}} & \multirow{2}{*}{1} \\

{} & {} &  $\mathcal{D}_{f}$ $\downarrow$ & 81.42 & 0 & 89.73 & \textbf{0} & {} & {}\\
\cline{2-9}
 & \multirow{2}{*}{ResNet9} &  $\mathcal{D}_{r}$ $\uparrow$ & 82.76 & 94.24 & 53.75 & \textbf{39.44} & \multirow{2}{*}{\textbf{143}} & \multirow{2}{*}{2} \\

{} & {} &  $\mathcal{D}_{f}$ $\downarrow$ & 87.11 & 0 & 49.65 & \textbf{0} & {} & {}\\
\hline

\multirow{4}{*}{MNIST} & \multirow{2}{*}{LeNet} &  $\mathcal{D}_{r}$ $\uparrow$ & 98.15 & 98.73 & 96.96 & \textbf{95.79} & \multirow{2}{*}{\textbf{1}} & \multirow{2}{*}{1} \\

{} & {} &  $\mathcal{D}_{f}$ $\downarrow$ & 99.59 & 0 & 99.37 & \textbf{0} & {} & {}\\
\cline{2-9}

 & \multirow{2}{*}{ResNet9} &  $\mathcal{D}_{r}$ $\uparrow$ & 98.57 & 98.83 & 12.32 & \textbf{94.57} & \multirow{2}{*}{\textbf{2}} & \multirow{2}{*}{3} \\

{} & {} &  $\mathcal{D}_{f}$ $\downarrow$ & 99.10 & 0 & 0 & 0 & {} & {} \\
\hline
\end{tabular}
\label{table:lenet}
\end{table*}


\section{Experiments}
We show the performance of the proposed methods for unlearning single and multiple classes in the zero-shot setting across multiple benchmark datasets. We use AllCNN~\cite{springenberg2014striving}, LeNet~\cite{lecun1998gradient} and ResNet9 (we build a smaller variant of ResNet~\cite{he2016deep}) models for zero-shot machine unlearning analysis on MNIST \cite{lecun1998gradient}, CIFAR-10~ \cite{krizhevsky2009learning} and SVHN \cite{netzer2011reading}.



\subsection{Experimental Settings}
The experiments are conducted on a NVIDIA Tesla-V100 (32GB) GPU. The original models and retrained models have been trained with a batch size of 256 for 40 epochs on CIFAR-10, 10 epochs on SVHN, and 10 epochs on MNIST. Without loss of generality, we unlearn class 0 in 1-class and classes 1 and 2 in 2-classes unlearning.\par

\textbf{Implementation details for error min-max noise method.} We optimize i.e., minimize or maximize a batch of noise (256 samples) for 400 steps with an initial learning rate of 0.1 for each class. If the loss doesn't improve, the learning rate is decreased by a factor of 0.5. On CIFAR 10, we use 2 impair steps~\cite{tarun2021fast} with a learning rate of 0.01. On SVHN, we use 3 impair steps with a learning rate of 0.001. On MNIST we use a single impair step with a learning rate of 0.01.\par

\textbf{Implementation details for gated knowledge transfer (GKT) method.} The generator and student are trained alternatively with 1 step for generator, and 10 steps for student. The KL temperature for loss function was set to 1 for MNIST and SVHN, and 0.5 for CIFAR-10. $\beta$ is set to 250. The learning rate for both student and generator for MNIST is set to 0.01. The learning rate for both student and generator for other datasets is set to 0.001. The training is performed for 4000 epochs and the threshold for filter $F$ is set to 0.01. We show the results of the model checkpoint with best performance on $D_r$ while still having $0\%$ accuracy on $D_f$.\par


\begin{table*}[]
    \centering
    \caption{A comparison between proposed zero-shot unlearning method with the existing unlearning methods.}
    \begin{tabular}{c|c|c|c|c|c}
    \hline
        Attributes & UNSIR~\cite{tarun2021fast} & Amnesiac~\cite{graves2021amnesiac}  & Bad Teacher~\cite{chundawat2022can}& Fisher~\cite{golatkar2020eternal}& \textbf{Our Method}\\
        \hline
        Class-level & \multirow{2}{*}{{\tikzcmark}} & \multirow{2}{*}{{\tikzcmark}} & \multirow{2}{*}{{\tikzcmark}} & \multirow{2}{*}{{\tikzcmark}} & \multirow{2}{*}{{\tikzcmark}}\\
        unlearning?  & & & & &\\
        \hline
        Random samples & \multirow{2}{*}{{\tikzxmark}} & \multirow{2}{*}{{\tikzcmark}} & \multirow{2}{*}{{\tikzcmark}} & \multirow{2}{*}{{\tikzcmark}} & \multirow{2}{*}{{\tikzxmark}}\\
        unlearning?  & & & & &\\
        \hline
        Zero-glance & \multirow{2}{*}{{\tikzcmark}} & {\multirow{2}{*}{\tikzxmark}} & \multirow{2}{*}{{\tikzxmark}} & \multirow{2}{*}{{\tikzxmark}} & \multirow{2}{*}{{\tikzcmark}}\\
        unlearning?  & & & & &\\
        \hline
        Zero training & \multirow{2}{*}{{\tikzxmark}} & \multirow{2}{*}{{\tikzxmark}} & \multirow{2}{*}{{\tikzxmark}} & \multirow{2}{*}{{\tikzxmark}} & \multirow{2}{*}{{\tikzcmark}}\\
        samples? & & & & &\\
        \hline
    \end{tabular}
\label{tab:metric_aspect_comparison}
\end{table*}

\begin{table*}[]
\footnotesize
\centering
\caption{Unlearning result comparison with recent state-of-the-art methods~\cite{chundawat2022can,graves2021amnesiac,tarun2021fast} (Not zero-shot).The comparisons are done for AllCNN+CIFAR-10. Our method performs quite well considering it has no access to the training dataset.}
\begin{tabular}{c|c|cc|cc}
\hline
\multirow{2}{*}{Method} & \multirow{2}{*}{Zero-shot?} &\multicolumn{2}{c}{1-class Unlearning} & \multicolumn{2}{c}{2-class Unlearning}\\
\cline{3-6}
&&$\mathcal{D}_{r}$ $\uparrow$ & $\mathcal{D}_{f}$ $\downarrow$ & $\mathcal{D}_{r}$ $\uparrow$ &$\mathcal{D}_{f}$ $\downarrow$ \\ 
\hline
\hline
Original Model& NA & 84.05 &87.49& 84.18 & 84.72\\
\hline
\hline
Retrain Method& NO & 85.72 & 0 & 86.30  & 0 \\
\hline
Bad Teacher~\cite{chundawat2022can}& NO & 83.71 & 5.56 & 84.11 & 7.81\\
\hline
Fisher~\cite{golatkar2020eternal}& NO & 7.61 & 0 & 8.57 & 0\\
\hline
Amnesiac~\cite{graves2021amnesiac}& NO & 83.47 & 0 & 82.85 & 0\\
\hline
Min-Max (ours)& YES & 20.48 & 5.11 & 29.59 & 7.59\\
\hline
GKT (ours)& YES & \textbf{81.97}& \textbf{0} & \textbf{81.70} & \textbf{0}\\
\hline
\end{tabular}
\label{table:non-zero-shot-baseline}
\end{table*}
\subsection{Evaluation Metrics} 
In the literature~\cite{tarun2021fast,golatkar2020eternal,golatkar2021mixed,graves2021amnesiac}, different metrics are employed to evaluate the unlearning methods. Usually, such metrics require both the training and testing data to measure the performance. For our zero-shot methods, the evaluation is done with the help of the training and testing sets even though such data is not used for the purpose of unlearning. To evaluate the robustness of the unlearning method, we use the following metrics:\par
\textit{Accuracy on the forget set $\mathcal{D}_{f}$ and the retain set $\mathcal{D}_{r}$} - The accuracy on $\mathcal{D}_{f}$ is desired to be closer to the retrained model as the intended behaviour of an unlearned model should be similar to that of the retrained model after unlearning. The accuracy on $\mathcal{D}_{r}$ is desired to be closer to the original model.\par

\textit{Anamnesis Index (AIN)}: As discussed in Section~\ref{AIN}, the \textit{AIN} should be close to 1. A margin ($\alpha$) of $5\%$, with a batch size of 256 was used for relearn time calculation. We use a learning rate of 0.1 and decrease the rate by a factor of 0.1 if accuracy does not improve on 2 successive epochs.\par
\textit{Model inversion attack}: The data obtained after inversion from the unlearned model should not contain any information about the forget class.\par
\textit{Membership inference attack}: The inference attack probability should be lower in the unlearned model in comparison to the original model for the data in the forget class.

\subsection{Results and Analysis}
We conduct several experiments and evaluate the single and multi-class unlearning performance in \textit{zero-shot setting} for the proposed baseline (Min-Max) method and gated knowledge transfer (GKT) methods. In Table~\ref{table:allcnn}, we show the results of AllCNN model over CIFAR-10, SVHN, and MNIST. Similarly, Table~\ref{table:lenet} gives the results for single-class unlearning in LeNet and ResNet9.

\subsubsection{\textbf{Baseline results.}} From Table~\ref{table:allcnn} and Table~\ref{table:lenet}, we observe that the baseline Min-Max method gives poor results in zero-shot unlearning setup. As this method requires optimizing the noise for both the forget and retain classes, the quality of noise samples determine the unlearning performance. The Min-Max method is not able to preserve the accuracy on $\mathcal{D}_{r}$. For example, in 1-class unlearning on AllCNN+CIFAR-10 (Table~\ref{table:allcnn}), it obtains 20.58\% accuracy on $\mathcal{D}_{r}$ in comparison to the 84.05\% accuracy of the original model. Similarly, the accuracy on $\mathcal{D}_{f}$ is 5.11\%. In 2-class unlearning on AllCNN+MNIST, only 10.96\% retain accuracy is obtained in comparison to the 98.17\% accuracy in the original model. Even when it obtains decent accuracy on $\mathcal{D}_{r}$ (Table \ref{table:allcnn}, SVHN, 1-class unlearning: 58.70\%), the accuracy on $\mathcal{D}_{f}$ is also high (50.23\%) which is not desirable. The low Anamnesis index (AIN) for this method indicates that the model quickly regains most of the information of $\mathcal{D}_{f}$. It means the unlearned model still remembers a lot of the $\mathcal{D}_{f}$ information. Similarly, the Min-Max method gives poor results for LeNet and ResNet9 as shown in Table~\ref{table:lenet}.\par

\subsubsection{\textbf{Results on GKT Vs Min-Max}}
The proposed GKT method achieves accuracy close to the original model on $\mathcal{D}_{r}$. For example, in 1-class unlearning on AllCNN+CIFAR-10 (refer Table \ref{table:allcnn}), GKT obtains 81.97\% accuracy on $\mathcal{D}_{r}$ which is quite close to the original models accuracy of 84.05\%. In 2-class unlearning on AllCNN+MNIST, the proposed GKT achieves 96.87\% accuracy on $\mathcal{D}_{r}$ in comparison to 98.17\% accuracy of the original model. In both cases, the accuracy on $\mathcal{D}_{f}$ is 0\% which quite good. The AIN score of the proposed GKT method in CIFAR-10+AllCNN and MNIST+AllCNN is 0.81 and 0.65, respectively. In both cases the AIN score is much closer to 1 in comparison to Min-Max method. This indicates that the unlearned model doesn't contain much information about the forget classes.\par

In Table~\ref{table:lenet}, the performance on LeNet and ResNet9 models are shown for 1-class unlearning. The proposed GKT method obtains 0\% accuracy on $\mathcal{D}_{f}$ across all the datasets. Furthermore, $\mathcal{D}_{r}$ accuracy is much closer to the original model in comparison to the Min-Max method. Although Min-max is consistent in terms of $\mathcal{D}_{r}$ and $\mathcal{D}_{f}$, it obtains significantly poor results in both of them. An ideal unlearning method should maintain the previous accuracy on $\mathcal{D}_{r}$ and give near zero accuracy on $\mathcal{D}_{f}$. The noise matrices generated through error-maximization (for forget class samples) and error-minimization (for retain class samples) in Min-Max ends up damaging the model for both the retain and forget data. Whereas, the GKT method is able to carefully refine the model weights through a teacher-student framework and a band-pass filter. Thus, overall, GKT gives better performance over Min-Max.\par

It is to be noted that the accuracy on $\mathcal{D}_{r}$ is not fully preserved by our methods. This is due to the zero-shot setting where the model can't observe any real data. In future, further improvements should be explored to obtain same $\mathcal{D}_{r}$ accuracy as the original model in zero-shot setting. We notice some values of AIN in Table~\ref{table:lenet} are anomalous. This is because the retrained model reached the target accuracy in just 1 epoch and therefore, the ratios are very high. The results on LeNet and ResNet9 are decent but not as consistent as AllCNN.

\begin{figure*}[t]
    \centering
    \includegraphics[width=0.25\textwidth]{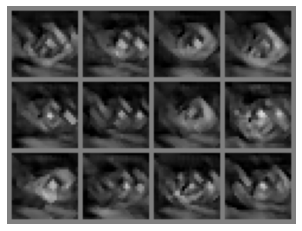}\hspace{-0.5em}
    \includegraphics[width=0.25\textwidth]{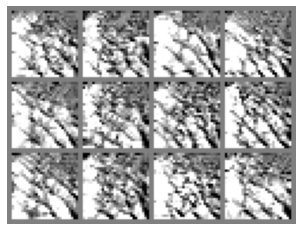}\hspace{-0.5em}
    \includegraphics[width=0.25\textwidth]{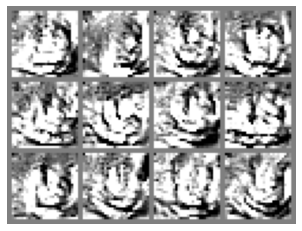}
\caption{Model inversion attack on AllCNN model for class-0. \textit{left:} Fully trained model  \textit{middle:} Model retrained without class-0 \textit{right:} Forget model obtained using the proposed GKT method. The images are shown for 12 independent attacks. The inversion attack fails in the proposed GKT method and the retrained model.}
\label{fig:Inversion-Attack}
\end{figure*}

\begin{figure*}[t]
\centering
    \centering
    \includegraphics[width=0.32\textwidth]{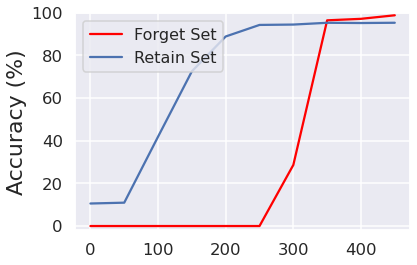} \hspace{-0.5em}
    \includegraphics[width=00.30\textwidth]{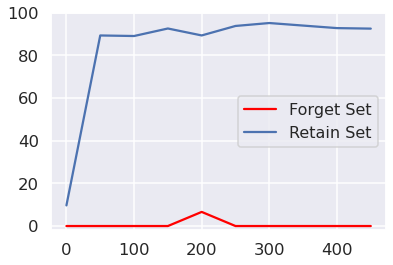} \hspace{-0.5em}
    \includegraphics[width=00.30\textwidth]{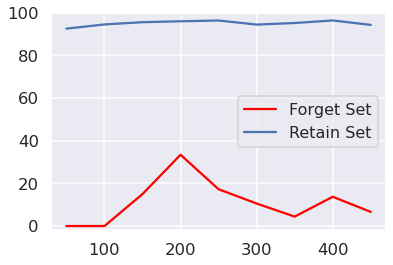} \hspace{-0.5em}
    \centering
    \includegraphics[width=0.32\textwidth]{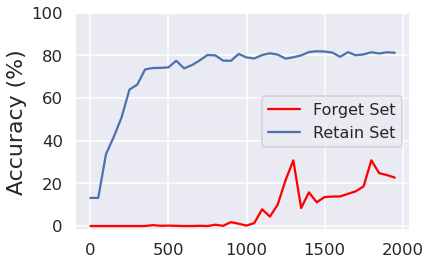} \hspace{-0.5em}
    \includegraphics[width=00.30\textwidth]{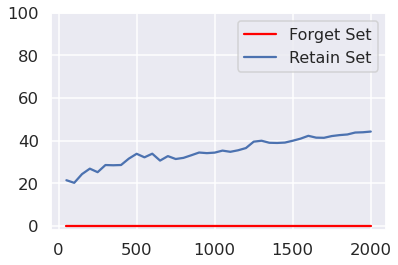} \hspace{-0.5em}
    \includegraphics[width=00.30\textwidth]{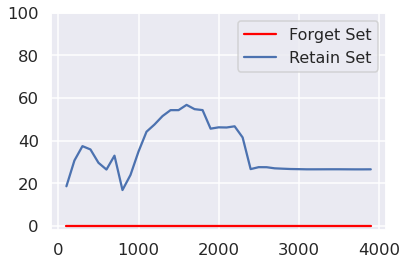} \hspace{-0.5em}
\bigskip
    \centering
    \includegraphics[width=0.32\textwidth]{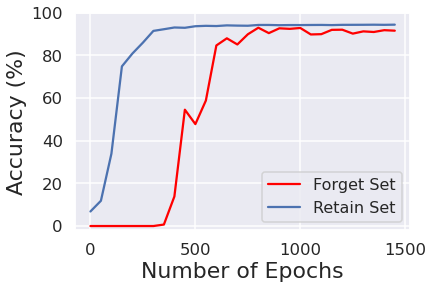} \hspace{-0.5em}
    \includegraphics[width=00.30\textwidth]{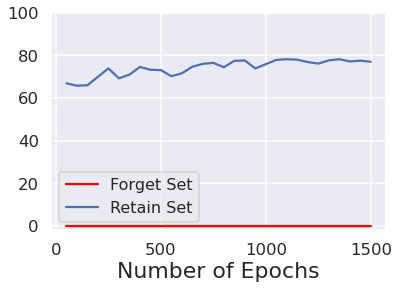} \hspace{-0.5em}
    \includegraphics[width=00.30\textwidth]{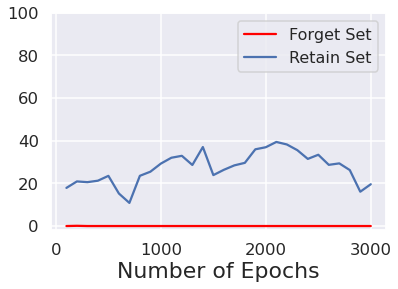} \hspace{-0.5em}
\caption{The GKT method progression with number of epochs in 1-class unlearning on MNIST (top), CIFAR-10 (middle), and SVHN (bottom). The models used are AllCNN (left), LeNet (middle), and ResNet9 (right). The graphs show how stopping after the right number of epochs is important. (Table~\ref{table:allcnn} and Table~\ref{table:lenet} present the results just before the accuracy on $\mathcal{D}_{f}$ shoots up.)}
\label{fig:gkt}
\end{figure*}

\subsubsection{\textbf{Comparison with the state-of-the-art unlearning methods}}
One major difference between our work and the existing state-of-the-art unlearning methods is that our method is zero-shot i.e., it does not require any access to training data. Whereas, the existing methods~\cite{chundawat2022can,graves2021amnesiac,golatkar2020eternal,tarun2021fast} need the training set data to perform unlearning. We show a comparison between different unlearning methods in Table~\ref{tab:metric_aspect_comparison} based on several attributes such as \textit{whether the method supports unlearning for class-level and a random set of samples}. Similarly, we compare what kind of data is required for the corresponding algorithm to function. Ours is the only method that can function without using any training data. We then compare our \textit{zero-shot} results with the \textit{non zero-shot} results in Table~\ref{table:non-zero-shot-baseline}. 
Our method lags behind~\cite{graves2021amnesiac} by 1.5\% in terms of $\mathcal{D}_{r}$ but it forgets $\mathcal{D}_{f}$ with equal effectiveness as~\cite{graves2021amnesiac}. The \textit{non zero-shot} methods give better results on the retain set but this is not a fair comparison because \textit{zero-shot} method can function even if training data is not available for unlearning. Whereas, the other methods can not function in the absence of training data.

\subsubsection{\textbf{Analysis}}
In Fig. 4, we show the progression in ~\textit{knowledge transfer} through the forget $\mathcal{D}_{f}$ and retain set $\mathcal{D}_{r}$ accuracy with increasing number of epochs. The GKT method uses a generator that maximizes the information gap between the teacher and student. Thus, it generates highly informative samples for the student to learn from. We use the band-pass filter to reject the samples generated corresponding to the forget set to be passed on to the student. After several iterations of training, most of the information gap corresponding to $\mathcal{D}_{r}$ will be bridged and we will be left with the information gap corresponding to $\mathcal{D}_{f}$ only. On the other hand, due to the adversarial setting, most of the generated samples will correspond to $\mathcal{D}_{f}$. Now the information gap lies in $\mathcal{D}_{f}$ samples only as other information has been already absorbed by the student. This means that the performance on $\mathcal{D}_{r}$ will stagnate and that on $\mathcal{D}_{f}$ will begin to rise. This behaviour can be observed in Fig. 4. For class-0 forgetting in AllCN+MNIST, we see a sharp rise around epoch $250$. If the performance tolerance for $\mathcal{D}_{f}$ is $0\%$, this is where we stop i.e., as soon as $\mathcal{D}_{f}$ accuracy becomes $>0\%$. In LeNet model unlearning, $\mathcal{D}_{f}$ accuracy stays at $0\%$. This means further training is possible and more performance on $\mathcal{D}_{r}$ can be recovered without affecting the performance on $\mathcal{D}_{f}$. In general, one should select the best checkpoint before the accuracy on $\mathcal{D}_{f}$ starts to rise. The checkpoints before that event contain information corresponding to $\mathcal{D}_{r}$ only and they should be picked as final parameters of an unlearned model.

\subsection{Privacy Attacks: Model Inversion Attack}
We check the data leakage in the the proposed methods using the state-of-the-art privacy attacks such as model inversion and membership inference attacks. We discuss the model inversion attack and robustness analysis below and the analysis for membership inference attack is given in the Supplementary document.

\subsubsection{\textbf{Threat Model}} 
In our analysis, we assume the adversary has white-box access to the current unlearned model but does not have access to the previous versions of the model. The adversary does not hold the information about the actual contents of each class. An attack is considered successful if the adversary is able to gather information about the representation of a class through model inversion. We use the model inversion attack presented in~\cite{graves2021amnesiac} which is a modified version of the attack in~\cite{fredrikson2015model}. A small amount of noise is added to the input vector initialized with zeros. The model is optimized with gradient descent using the loss computed with respect to this input and the target class. An image processing step is executed after every $n$ steps of gradient descent which helps in the recognition of the generated images. The whole process is repeated for some number of epochs to obtain the final inverted images.\par

\subsubsection{\textbf{Robustness Analysis}}
The model inversion attacks were performed on the AllCNN model trained on MNIST. Without loss of generality, the attacks were performed before and after forgetting class-0 from the model. Fig.~\ref{fig:Inversion-Attack} shows the inversion attack results on fully trained model, model trained without class-0 i.e., \textit{retrained model}, and forget model obtained using the proposed GKT method. The inversion attacks on fully trained model shows circle like patterns among inverted images hinting at possible data leakage. However, the patterns are completely random for gold model and the proposed forget model. The model inversion attacks are not able to extract any information from our forget model. This indicates that our method is successful at removing information regarding class-\textit{0} and is robust to the model inversion attacks.

\subsection{Ablation Study} 
We conduct several ablative experiments to further analyze the proposed method. Particularly, several variations are carried out on the GKT method to observe the change in its effectiveness. We discuss the results of the following ablation studies.


\subsubsection{\textbf{Band-Pass filter ablation}} 
\begin{figure*}
\centering
    \includegraphics[width=0.34\textwidth]{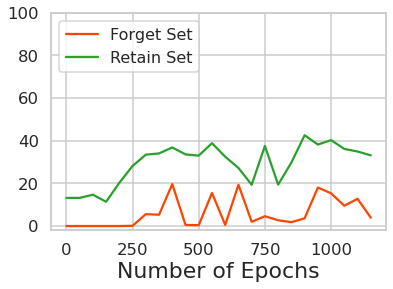}\hspace{-0.5em}
    \includegraphics[width=0.34\textwidth]{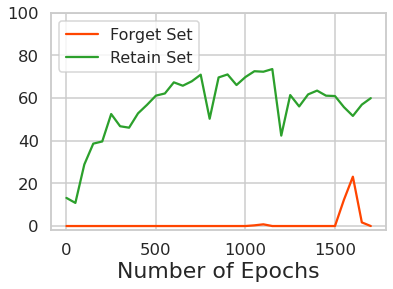}\hspace{-0.5em}
    \includegraphics[width=0.34\textwidth]{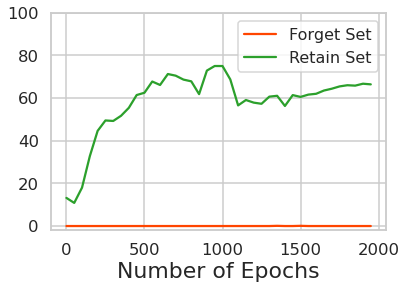}\hspace{-0.5em}
    \includegraphics[width=0.34\textwidth]{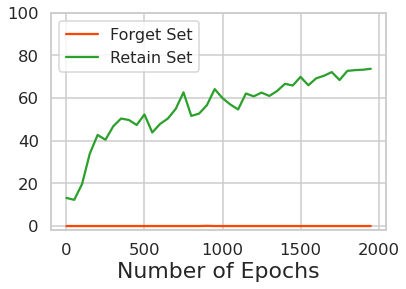}\hspace{-0.5em}
\caption{Effect of varying the threshold in band pass filter. From top-left to bottom-right, the thresholds are 0.5, 0.1, 0.01, and 0.001, respectively. As visible, thresholds more than random probability ($1/10 = 0.1$) are ineffective at stopping information flow corresponding to $D_f$.}
\label{fig:threshold}
\end{figure*}

The effect of~\textit{varying the filter's threshold} ($\epsilon$) on the quality of knowledge transfer in GKT method is depicted in Fig.~\ref{fig:threshold}. The threshold values of 0.001 and 0.01 work quite well in our experiments. The default value of 0.01 converges faster. Using a value of 0.1 makes the generator crash around epoch 1750. It means after this none of the samples produced by the generator pass the filter and the student doesn't train anymore. The threshold of 0.5 leads to the worst outcome. It is neither stable nor does it reach anywhere near to the target performance throughout its training. This is expected as even a randomly initialized model will give an average probability of 0.1 (as there are 10 classes) corresponding to the forget class and any threshold above this shouldn't work.

\begin{figure*}[]
\captionsetup{skip=0pt}
    \centering
    \includegraphics[width=0.32\textwidth]{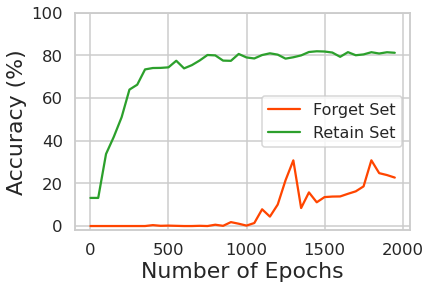} \includegraphics[width=0.30\textwidth]{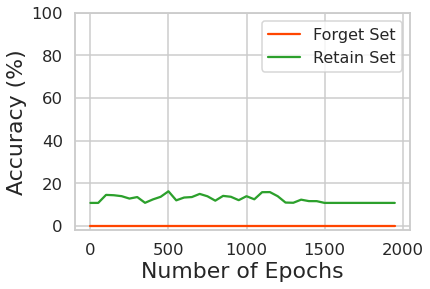}
    \includegraphics[width=0.30\textwidth]{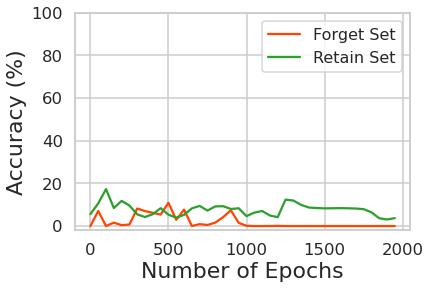}
\caption{Ablation study on the loss function of the proposed GKT method. \textit{Left:} Default experiment settings. \textit{Middle:} When generator has a attention difference term in its loss. \textit{Right:} When JS-divergence is used instead of KL-divergence in both student and generator loss.}
\label{fig:ablation,att,js}
\end{figure*}

\subsubsection{\textbf{Loss functions}}
We make the following variations to the loss function: (i)~\textit{add the attention difference to the generator's loss}, (ii)~\textit{replace KL-divergence in the training procedure with Jenson-Shannon (JS)-divergence}. The results with these 2 losses are shown in~Fig.~\ref{fig:ablation,att,js}. Adding attention difference to the generator's loss makes it too easy for the generator to fool the student. Therefore the student doesn't learn anything and same can be observed in the middle graph in Fig.~\ref{fig:ablation,att,js}. Similarly, the JS-divergence also doesn't have much impact on the learning capacity of the student. It instead decreases the effectiveness of GKT compared to the default KL-Divergence.

\subsection{Limitations}
To the best of our knowledge, this paper presents the first solution for zero shot unlearning. However, there are some limitations such as the proposed method is not as effective in very large models. To demonstrate the effectiveness of data removal from the deep learning model, a variety of widely accepted metrics have been evaluated like performance on $\mathcal{D}_{r}$, $\mathcal{D}_{f}$, relearn time, and privacy attacks, including a newly proposed AIN metric. A more formal guarantee of unlearning might be desired in highly privacy-sensitive applications. Currently our method supports class-level unlearning. Forgetting a random cohort of data, or a subset of a class is beyond the scope of this work. For example, if there was a class for vehicles, forgetting just a specific vehicle like a car of a particular company can be considered as a subclass removal problem and that is not supported by our method.

%% file: sec/5_conclusions.tex
\section{Conclusion}
We introduced the problem of~\textit{zero-shot machine unlearning} that opens up new challenges for unlearning in a stricter setting. It also offers a framework to keep up with the modern privacy regulations that include the provision for \textit{right to be forgotten}. We present two novel \textit{zero-shot} approaches to this problem: error minimization-maximization noise and gated knowledge transfer. The gated knowledge transfer approach proves to be highly effective in case of both single and multiple-class unlearning. Several ablative studies are presented to gain insight into the working principles of the proposed method. We introduce a new evaluation metric, Anamnesis Index (AIN), to measure the quality of unlearning in the ML models. The experiments are conducted on  benchmark datasets and the results are quite promising in zero-shot setting. Protection against information leak is evaluated under privacy attacks such as model inversion attack and membership inference attack. Our work gives a new direction of research in a more stringent, but realistic setting for machine unlearning. A possible direction of future research can be to try zero-shot unlearning on deeper networks and large datasets. Another future direction could explore obtaining a theoretical bound on the amount of information remaining in the model after forgetting. 


\section*{Acknowledgment}
This research/project is supported by the National Research Foundation, Singapore under its Strategic Capability Research Centres Funding Initiative. Any opinions, findings and conclusions or recommendations expressed in this material are those of the author(s) and do not reflect the views of National Research Foundation, Singapore.

%% file: sec/X_supplementary.tex
\newpage
\onecolumn
\appendix

\begin{algorithm}[]
\caption{Noise Minimization-Maximization}
\label{alg:MINMAX}
\begin{algorithmic}[1]
\State M(.;$\phi$) (Fully Trained Model)
\State $C_f \gets$ forget classes
\State $C_r \gets$ retain classes
\State Noises $\gets [\;]$
\For{$i \in C_f$} 
\State Training Noise for forget classes
\State $\mathcal{N}_f \gets N(0, I)$ (Initialize Noise)
\For{1,2,....,$n_{C_f}$}
    \State $L_N \gets -\mathcal{L}(M(z;\phi), i) + \lambda||z||_2$
    \State $\mathcal{N}_f \gets \mathcal{N}_f - \eta \frac{\partial L_N}{\partial \mathcal{N}_f}$
\EndFor
\State Noises.append($\mathcal{N}_f$, $i$)
\EndFor

\For{$i \in C_r$} (Training Noise for retain classes)
\State $\mathcal{N}_r \gets N(0, I)$ (Initialize Noise)
\For{1,2,....,$n_{C_r}$}
    \State $L_N \gets \mathcal{L}(M(z;\phi), i)$
    \State $\mathcal{N}_r \gets \mathcal{N}_r - \eta \frac{\partial L_N}{\partial \mathcal{N}_r}$
\EndFor
\State Noises.append($\mathcal{N}_r$, $i$)
\EndFor
\State dataset $\gets Noises$
\State shuffle(dataset)
\For{1,2....$n_{epochs}$}
\For{batch in dataset} \Comment{Perform the unlearning step.}
\State input, labels $\gets$ batch
\State predictions $\gets$ $M(input;\phi)$
\State $L_M \gets \mathcal{L}$(predictions, labels)
\State $\phi \gets \phi - \eta \frac{\partial L_M}{\partial \phi}$
\EndFor
\EndFor
\end{algorithmic}
\end{algorithm}

\begin{algorithm}[]
\caption{Gated Knowledge Transfer}
\label{alg:GKT}
\begin{algorithmic}[1]
\State T(.) (Fully Trained Teacher Network)
\State S(.;$ \theta$)  (Randomly Initialized Student)
\State G(.;$ \phi$)  (Randomly Initialized Generator)
\State $C_f \gets$ forget classes
\State $threshold \gets { \epsilon}$
\For{1,2,....,$n_{epochs}$}
\State $z \gets N(0, I)$
\For{1,2,....,$n_G$}
    \State $x_p \gets G(z; \phi)$
    \State $x_p \gets F(x_p, C_f, threshold)$ (Applying Filter)
    \State $L_G \gets -D_{KL}(T(x_p) || S(x_p))$
    \State $\phi \gets \phi - \eta \frac{\partial L_G}{\partial \phi}$
\EndFor

\For{1,2,....,$n_S$}
    \State $x_p \gets G(z; \phi)$
    \State $x_p \gets F(x_p, f_{cls}, threshold)$ (Applying Filter)
    \State $L_S \gets D_{KL}(T(x_p) || S(x_p)) + \beta L_{at}$
    \State $\theta \gets \theta - \eta \frac{\partial L_S}{\partial \theta}$
\EndFor
\EndFor
\end{algorithmic}
\end{algorithm}

\begin{figure*}[t]
    \centering
    \includegraphics[width=0.4\textwidth]{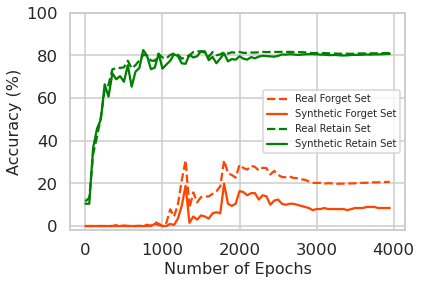}
    \includegraphics[width=0.4\textwidth]{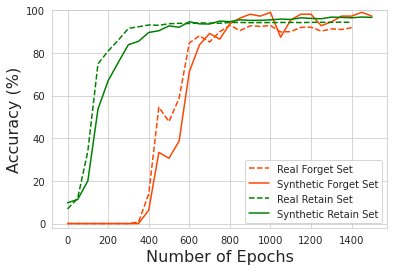}
    \caption{GKT progress comparison between real and synthetic data. The synthetic test images were generated using DeepInversion.~\textit{left:} The progression of unlearning (class-0) on synthetic CIFAR-10 test data and real CIFAR-10 test data on AllCNN.\textit{right:} The progression of unlearning (class-0) on synthetic SVHN test data and real SVHN test data on AllCNN. The progression on synthetic data is highly correlated and similar to progression on real data and it can be used to decide stopping criteria in complete absence of test data.}
    \label{fig:synthetic}
\end{figure*}

\begin{figure*}[t]
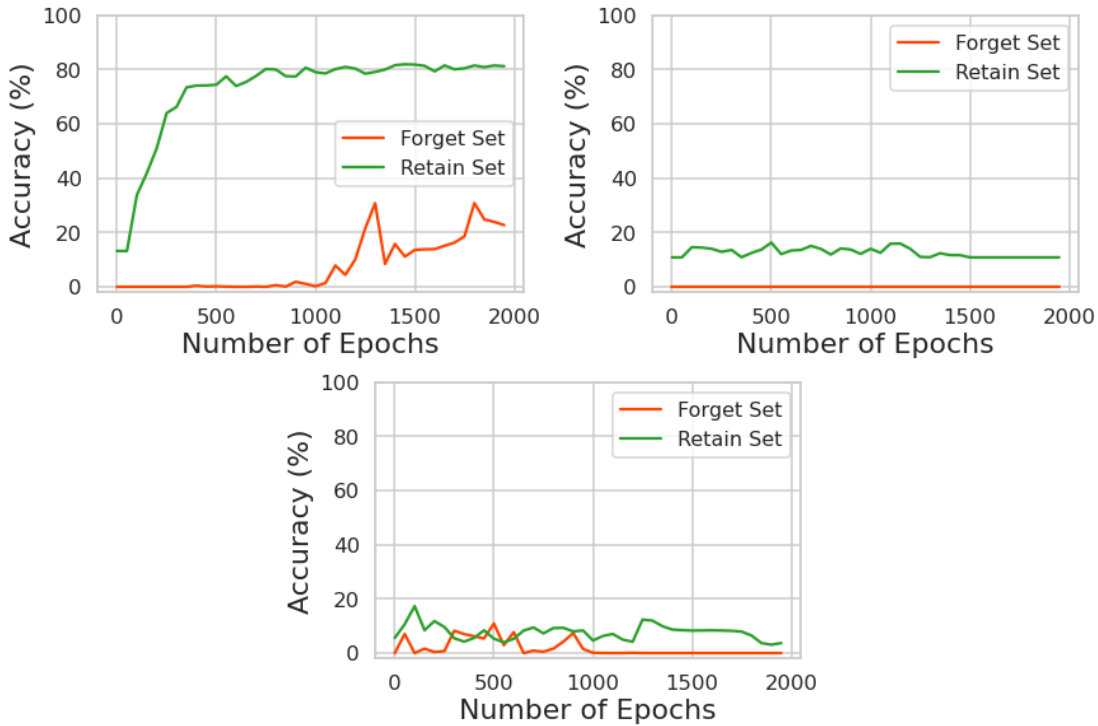

\captionsetup{skip=0pt}
\centering
    \centering
    \includegraphics[width=0.4\textwidth]{fig/CIFAR_AllCNN2.png}
    \includegraphics[width=0.4\textwidth]{fig/Attention_Difference.png}
    \includegraphics[width=0.4\textwidth]{fig/JS-Divergence.png}
\caption{Ablation study on the proposed GKT method. \textit{Top:} Default experiment settings. \textit{Middle:} When generator also has a attention difference term in its loss. \textit{Bottom:} When JS-divergence is used instead of KL-divergence in both student and generator loss.}
\label{fig:ablation,att,js}
\end{figure*}

\subsection{Algorithms}
Algorithm \ref{alg:MINMAX} describes the Noise Minimization-Maximization method proposed in Section IV.B. Similarly, Algorithm \ref{alg:GKT} describes the Gated Knowledge Transfer method proposed in Section IV.C. The generator architectures used in the GKT method is given in Table~\ref{table:gan_arch}.

\subsection{Additional Ablation Study}
\subsubsection{Stopping Criteria}
In our experiments, we have used the performance on the retain set and forget set to evaluate the progress of GKT and also to decide when to stop the training. However, in a truly \textit{absolute zero-shot} scenario we will not even have access to these forget and retain sets even for evaluation purposes. In such cases, methods like Deep Inversion
\footnote{Yin et al., “Dreaming to distill: Data-free knowledge transfer via deepinversion,” in Proceedings of the IEEE/CVF Conference
on Computer Vision and Pattern Recognition, 2020, pp. 8715–8724} can be used to obtain synthetic samples which may act as a proxy to the real data. Figure~\ref{fig:synthetic} shows how the performance on the real retain set, synthetic retain set, real forget set, and synthetic forget set progress with increasing number of steps of distillation. From Figure~\ref{fig:synthetic}, it is clearly visible that the performance on the synthetic set is highly indicative of how the performance might have been on the real data. The performance on the synthetic retain set is nearly identical to that of the performance on the real retain set. Similarly, the performance on the synthetic forget set shoots up at nearly the same epoch as that on the real forget set. This means that the performance on the synthetic forget set can used to  decide when to stop even in an \textit{absolute zero-shot setting}.

\begin{table}[]
\footnotesize
\centering
\caption{The architecture detail of the generator used in GKT method}
\begin{tabular}{c|c}
\hline
\textbf{Layer} & \textbf{Dimensions/Parameters}\\
\hline
Linear & (z\_dim, 128*64)\\
\hline
Reshape & (128, 8, 8)\\
\hline
BatchNormalization & -\\
\hline
Upsample & factor=2\\
\hline
Convolution & (128, 128, 3)\\
\hline
BatchNormalization & -\\
\hline
LeakyReLU & negative\_slope = 0.2\\
\hline
Upsample & factor=2\\
\hline
Convolution & (128, 64, 3)\\
\hline
BatchNormalization & -\\
\hline
LeakyReLU & negative\_slope = 0.2\\
\hline
Convolution 2D & (64, 3, 3)\\
\hline
BatchNormalization & -\\
\hline
\end{tabular}
\label{table:gan_arch}
\end{table}

\subsubsection{Adding Attention Difference to the Loss}
Figure~\ref{fig:ablation,att,js} shows the results when we \textit{add the attention difference to the generator's loss} and when we replace KL-divergence in the training procedure with \textit{JS (Jenson-Shannon)-divergence}. Adding attention difference to the generator's loss makes it too easy for the generator to fool the student, and the student doesn't learn. Also, if JS-divergence is used, the method doesn't have its intended effect and the student again doesn't learn.\par

\subsection{Individually Unlearning Every Class}
In addition to 1-class and 2-class unlearning presented in the main paper, we also show unlearning for each individual class (class 0 to 9) on AllCNN+CIFAR10. We then compare the results with the retrained model as given in Table~\ref{table:unlearning_all_classes}. Our zero-shot method gives same performance as the retrain model in the forget set (\textbf{$0\%$}). However, the difference in the performance on the retain set varies from negligible ($0.09\%$, class 4) to substantial ($10.53\%$, class 3 and $10.63\%$, class 5). The difficulty in unlearning individual class is different. This is due to the different disentanglement of each class representation in the model. The more disentangled the representation is of a class, the more retain accuracy can be maintained. Similarly, more entangled a class's representation is, the more difficult to obtain good retain accuracy, without affecting the forget set accuracy. The average deviation for the retain set is \textbf{$5.22\%$}.

\begin{table*}[]
\footnotesize
\centering
\caption{GKT method performance after every individual class unlearning in AllCNN+CIFAR10.}
\begin{tabular}{c|c|cccccccccc}
\hline

\multirow{2}{*}{Method} & \multirow{2}{*}{Accuracy} & \multicolumn{10}{|c}{Unlearning Class}\\
\cline{3-12}
{} & {} &  0 & 1 & 2 & 3 & 4 & 5 & 6 & 7 & 8 & 9\\
\hline
\multirow{2}{*}{GKT} & $\mathcal{D}_{f}$ $\downarrow$  & 0 & 0 & 0 & 0 & 0 & 0 & 0 & 0 & 0 & 0\\
{} & $\mathcal{D}_{r}$ $\uparrow$  & 81.97 & 77.48 & 79.94 & 77.63 & 84.44 & 76.91 & 77.92 & 82.11 & 83.01 & 82.41\\
\hline
\multirow{2}{*}{Retraining} & $\mathcal{D}_{f}$ $\downarrow$  & 0 & 0 & 0 & 0 & 0 & 0 & 0 & 0 & 0 & 0\\
{} & $\mathcal{D}_{r}$ $\uparrow$  & 85.72 & 84.41 & 86.47 & 88.16 & 84.53 & 87.54 & 85.42 & 84.38 & 84.99 & 84.36\\
\hline
\multirow{2}{*}{Difference} & $\mathcal{D}_{f}$  & 0 & 0 & 0 & 0 & 0 & 0 & 0 & 0 & 0 & 0\\
{} & $\mathcal{D}_{r}$  & 3.75 & 6.93 & 6.53 & 10.53 & 0.09 & 10.63 & 7.50 & 2.27 & 1.98 & 1.95\\
\hline
\end{tabular}
\label{table:unlearning_all_classes}
\end{table*}

\begin{figure*}
    \centering
    \includegraphics[width=0.6\textwidth]{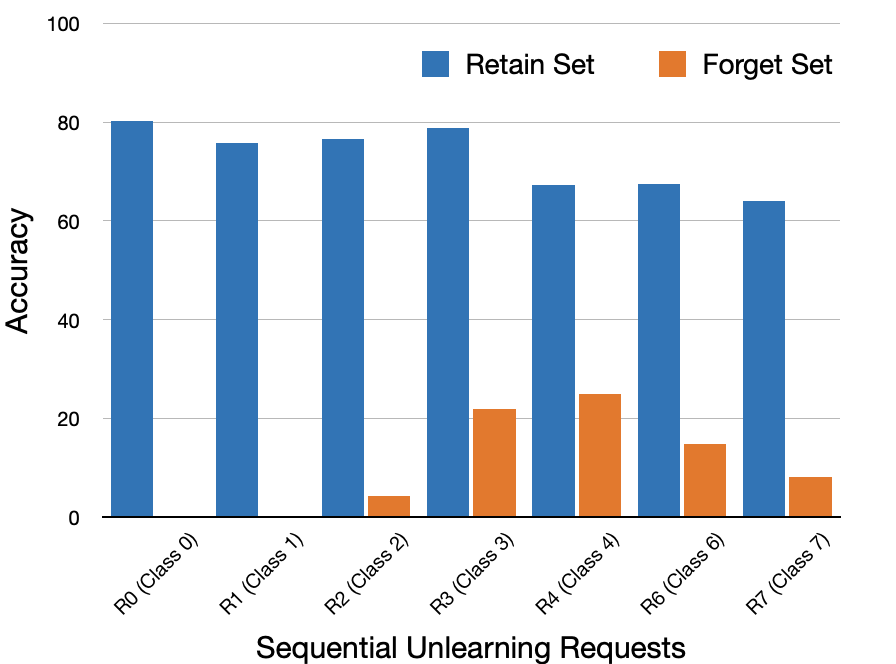}
    \caption{Sequential unlearning results in CIFAR-10+AllCNN. We show results on 7 sequential unlearning requests.}
    \label{fig:sequential}
\end{figure*}

\begin{table*}[t]
\footnotesize
\centering
\caption{Membership inference attack probability on various subsets of the complete data. The fully trained model, retrained/gold model, and GKT model are used for analysis with AllCNN+CIFAR-10. The membership inference attack fails on our model. It gives an attack probability of 1 for the train, test, forget, and retain sets. In the test set, the ideal attack probability should be \textit{zero}. However, the attack is giving an attack probability of 1 for the proposed GKT method. This indicates the inference attack is not consistent enough for checking the privacy leak.}
\begin{tabular}{c|cccc}
\hline
\multirow{2}{*}{} & \multicolumn{4}{c}{Membership Inference Attack Probability}\\
\cline{2-5}
{} & Train Set & Test Set & Retain Set & Forget Set\\
\hline
{Ideal Score} & 1 & 0 & 1 & NA\\
\hline
Fully Trained Model & 0.998 & 0.791 & 0.998 & 0.998\\
\hline
Retrained (Gold) Model & 0.939 & 0.748 & 0.998 & 0.403\\
\hline
GKT Model & 1 & 1 & 1 & 1\\
\hline
\end{tabular}
\label{table:inference-attack}
\end{table*}
\subsection{Sequential Unlearning}
In real world scenario, the deletion or forgetting requests may come at different instances. Therefore, it is important to validate the sequential unlearning performance of a method. Fig.~\ref{fig:sequential} shows the effect of sequentially unlearning 7 classes one after another on CIFAR-10+AllCNN. The model first unlearns class-0. This model is used to unlearn class-2 and the subsequent model then unlearns class-3, class-4, and so on. We notice that after $4^th$ sequential class unlearning, the performance on both retain and forget set begin to marginally dwindle. Overall, this degradation is gradual. For a 10-class model, after the $6^{th}$/$7^{th}$ class unlearning, it is better to retrain the model rather than unlearning more classes. The trade-off between the cost of unlearning and retraining must be investigated while making such decisions.

\subsection{Membership Inference Attack}
\subsubsection{Threat Model}
the adversary attempts to learn about the presence of a particular data point or a set of data points in the training set. The leakage of the membership information poses another kind of privacy threat. For example, an adversary may exploit the knowledge of an individual's data being present in a dataset that was used to train a model to study a certain disease. This could reveal the private medical history of that patient. We use the membership inference attack presented in~\cite{golatkar2020forgetting} to evaluate privacy leakage of the forget class of data. The attack is formulated as a binary classification problem, with the train set being label 1, and test set being label 0. The input is the entropy of the output probabilities of the model on which the attack is being performed. Membership inference attacks rely on the confidence of the models on a particular data points. A higher confidence indicates that the model has possibly seen the particular data point in training, whereas a more random prediction means the opposite.

\subsubsection{Robustness Analysis}
Due to the strict \textit{zero-shot setting} in our work, the proposed GKT method does not observe any data points while obtaining the unlearning model. The GKT uses the pseudo samples generated by the generator that contain the highest information for the student with respect to the teacher. The most informative samples are the ones with very high confidence on a few classes. It forces the model to always give output with high confidence. Thus, the inference attack probability for any query data point almost always tends to 1. 
Although, it might seem like there is privacy leakage in the model but this is not the case. We observe in Table~\ref{table:inference-attack}, for the unlearned model the inference attack probability is close to 1 for all types of data i.e., train data, retain data, forget data, and  even for test data. It is uniformly giving the same output for any kind of data. This suggests that the very nature of the zero-shot method makes it impossible for an membership inference attack to distinctly tell the difference between the data which is present and data which is not present in the model.~Rezaei and Liu~\cite{rezaei2021difficulty} discussed about several issues related to membership inference attacks and demonstrate their unreliable nature in numerous setups including deep learning models. Therefore, it is not surprising to see the failure of membership inference attacks in our \textit{zero-shot unlearning setting}. A good membership attack that is uniformly applicable
to different unlearning methods would greatly help the machine unlearning problem. This however, requires further investigation which is beyond the scope of this paper. 